\documentclass[journal,10pt]{IEEEtran}
\IEEEoverridecommandlockouts
\usepackage[style=ieee, urldate=comp]{biblatex}
\renewcommand*{\bibfont}{\footnotesize}

\usepackage{amsmath,amssymb,amsfonts}
\usepackage{algorithmic}
\usepackage{graphicx}
\usepackage{textcomp}
\usepackage{subfig}
\usepackage{array}
\usepackage{soul}
\usepackage{tabularx}
\usepackage{booktabs}
\usepackage{multirow}
\usepackage{multicol}
\usepackage{url}
\usepackage{slashbox}
\usepackage{siunitx}
\usepackage[table]{xcolor}
\usepackage{xspace}

\newcommand{\etal}{\textit{et al.}\xspace}

\newcolumntype{C}{>{\centering\arraybackslash}X}%

\begin{document}
\title{Detecting Intentional AIS Shutdown in Open Sea Maritime Surveillance Using Self-Supervised Deep Learning}

\author{Pierre Bernab\'{e}\IEEEauthorrefmark{1}\IEEEauthorrefmark{2},
Arnaud Gotlieb\IEEEauthorrefmark{1},
Bruno Legeard\IEEEauthorrefmark{2},
Dusica Marijan\IEEEauthorrefmark{1},
Frank Olaf Sem-Jacobsen\IEEEauthorrefmark{3} and
Helge Spieker\IEEEauthorrefmark{1}
\thanks{Manuscript received dd.mm.YYYY. This work was supported by the Norwegian Research Council (RCN) TSAR project under contract 287893 and benefited from the experimental infrastructure to explore exascale calculus (eX3), which RCN funds under contract 270053.
Satellite AIS data used for model development and testing has been made available courteously by its owner, the Norwegian Coastal Administration (Kystverket).}%
\thanks{\IEEEauthorrefmark{1}Simula Research Laboratory,
Oslo, Norway,
Email: \{pierbernabe,\allowbreak{}arnaud,\allowbreak{}dusica,\allowbreak{}helge\}@simula.no}%
\thanks{\IEEEauthorrefmark{2}Institut FEMTO-ST, Université de Bourgogne Franche-Comté,
Besançon, France, Email: bruno.legeard@femto-st.fr} %
\thanks{\IEEEauthorrefmark{3}Statsat AS, Oslo, Norway, Email: Frank.Sem-Jacobsen@statsat.no}%
}

\maketitle

\begin{abstract}  %
In maritime traffic surveillance, detecting illegal activities, such as illegal fishing or transshipment of illicit products is a crucial task of the coastal administration. In the open sea, one has to rely on  
Automatic Identification System (AIS) message transmitted by on-board transponders, which are captured by surveillance satellites.  
However, insincere vessels often intentionally shut down their AIS transponders to hide illegal activities. 
In the open sea, it is very challenging to differentiate intentional AIS shutdowns from missing reception due to protocol limitations, bad weather conditions or restricting satellite positions. 
This paper presents a novel approach for the detection of abnormal AIS missing reception based on self-supervised deep learning techniques and transformer models. Using historical data, the trained model predicts if a message should be received in the upcoming minute or not. Afterwards, the model reports on detected anomalies by comparing the prediction with what actually happens.
Our method can process AIS messages in real-time, in particular, more than \num{500} Millions AIS messages per month, corresponding to the trajectories of more than \num{60000} ships. The method is evaluated on 1-year of real-world data coming from four Norwegian surveillance satellites. 
Using related research results, we validated our method by rediscovering already detected intentional AIS shutdowns.
\end{abstract}


\section{Introduction}

\IEEEPARstart{V}{essel} Traffic Service (VTS) aims at monitoring and controlling the activities of vessels in dedicated maritime areas. The general objectives of VTS include identifying and guiding ships, helping vessels to prevent collisions, launching rescue missions at sea, or more generally, regulating the maritime traffic. In addition, modern VTS systems also support the detection of illegal activities such as piracy, fishing in protected zones, intrusion into economic exclusion zones, transshipment of narcotics, degassing at sea, etc. Most of the time, the detection of these illegal activities rely solely on the visual observation of vessels, manual analysis of collected data, and coastal administration officers' intuition based on their long-term experience~\cite{riveiro_maritime_2018}.

Among the sources of maritime surveillance information, Automatic Identification System (AIS) messages play an essential role. At sea, passenger ships or ships of sufficient tonnage must transmit their identity, their position, their direction and speed, and additional information up to every two seconds~\cite{international-maritime_organization-revised-2015}.
These messages are captured by various means, such as beacons at sea, coastal base stations, and satellites dedicated to observing maritime traffic. However, it happens that the transmission of AIS messages is absent for some vessels. 
It can be due to a failure of the AIS transponder, but it may also correspond to its intentional shutdown.

Actually, shutting down the AIS transmission is a simple action used by (a few) vessel captains to silently perform illegal actions at sea. For example, in order to fish in a prohibited area located in the open sea, a vessel can stay invisible by intentionally shutting down the AIS transponder~\cite{malarky-avoiding-nodate} during fishing. For instance, Oceana \cite{mustain_oceana_2021} reports that, from 2018 to 2021, it's no less than \num{600000} hours by \num{800} fishing ships that occurred in an illegal fishing zone close to the Argentina's national waters by using this technique.
Noticeably, Agnew {\it et al.} \cite{agnew-estimating-2009} estimate that a total loss between \$10 billion and \$23.5 billion can be imputed annually to illegal and unreported fishing activities worldwide, which shows that this problem has a huge economic impact. 
Desai and Shambaugh \cite{desai-measuring-2021} further emphasize the negative impact of illegal fishing on local fishing industries and its destruction of the corresponding ecosystem and the environment.

The problem is considered serious by the coastal administration all around the world and for prohibited maritime areas close to the coast (typically within a 20km-zone from the coast), different sources of information (e.g., radars, human visual control from ships or land stations) can be efficiently combined to automatically detect illegal activities.
However, when vessels transit at long distance from any coast in the open sea, the maritime surveillance can only rely on satellite AIS data. For instance, in the past ten years, the Norwegian Coastal Administration has been using satellites to capture AIS messages outside the areas covered by the base stations. 
Statsat AS, the company in charge of managing these satellites on behalf of the Norwegian government, drives the AISSat-\{1, 2\}\footnote{https://eoportal.org/web/eoportal/satellite-missions/a/aissat-1-2} and NorSat-\{1, 2, 3\}\footnote{https://eoportal.org/web/eoportal/satellite-missions/n/norsat-1-2} satellites which supervise AIS transmissions. 
These satellites are inclined at 97° and are positioned in Sun-synchronous polar orbit at an orbital height of 600-650 km, such that they shift to cover all latitudes with each rotation around the earth.

While these satellites ensure a broad coverage of the earth, it happens that some AIS messages remain unseen due to weather perturbations or reception conflicts. Also, the selected polar orbit leads to a better coverage of the poles rather than the equator. 
Hence, to human eyes, while reconstructing the trajectory of a single ship, it is common to have missing AIS messages on the trajectory, but for a given ship trajectory, distinguishing missing AIS messages reception due to an AIS transponder shutdown from acceptable causes is almost impossible. 
In order to distinguish both cases, we refer to \textit{abnormal missing AIS message reception} for the former case and as \textit{ordinary missing AIS message reception} for the latter.

Even though the number of messages that can be processed by the satellites is artificially limited, the volume of collected AIS messages is still considerable. In a single day, the Norwegian satellites can collect more than 4 million AIS messages (e.g., on 15/01/2020, there were 4,862,628 messages).
Due to the high volume, only a restricted proportion of AIS messages can actually be manually explored. To the best of our knowledge, there exists no dataset annotated by operators that differentiate intentional AIS shutdown from ordinary missing AIS message reception. It is therefore not possible to train models to directly detect these issues. For any alerting system to be actionable (i.e. sending out assets to investigate) the system needs to detect intentional AIS shutdown in near real-time. Prompt delivery of alerts ensures that the coastal administration can follow up on the alert while the illegal activity is performed. The large datasets, seemingly unpredictable nature of data gathering, and timeliness requirements, make it very challenging to detect intentional AIS shutdowns in the open sea and to propose an alerting system that can work in near real-time.

This paper proposes a deep learning-based approach to detect intentional AIS shutdowns in open sea. Our method uses self-supervision techniques to cope with the problem related to distinguishing ordinary from abnormal missing AIS message reception (which are witnesses of likely intentional AIS shutdown). 
Self-supervision means to extract pseudo-labels from the unlabelled data set~\cite{ericsson_self-supervised_2022}.
Our method is generally an unsupervised task, but as pseudo-labels can be derived from the dataset itself using self-supervision techniques, it allows us to eventually train the model in a supervised manner.

By training a multi-layer neural network model with auto-labelled data that learn the missing AIS messages' normality, we design a model that can process AIS messages in real-time and alert the coastal administration on suspicious ship trajectories containing possible intentional AIS shutdowns. More precisely, the results presented in this article revolve around three distinct contributions:

\begin{enumerate}
    \item We provide a self-supervised deep-learning methodology to detect abnormal missing AIS reception in the open sea. Our methodology is based on unlabelled world-covering satellite-collected AIS data, which has not been annotated by operators; 
    \item 
    Our methodology is designed for live surveillance of vessels. It exploits raw monthly data containing up to 500 million vessel messages corresponding to more than 60,000 unique vessels. Our DL model can be exploited in real-time to ease the decision-making process of the coastal administration;
    \item We demonstrate through experimental results the effectiveness of our methodology, as well as its robustness in varying scenarios and configurations. In particular, the use of the state-of-the-art \textit{transformer} deep learning model architecture benefits us with greater precision in the data analysis. 

\end{enumerate}

The rest of this paper is organized as follows:  
Section~\ref{sec:related_work} reviews previous work on the usage of Machine Learning in the maritime surveillance. 
We introduce our methodology with a general overview in Section~\ref{sec:overview}, before detailing how raw data are handled in Section~\ref{sec:data}. Section~\ref{sec:model} presents our methodology in depth as well as the architecture of our deep learning model and its training regime. In Section~\ref{sec:results}, we evaluate the methodology in experiments along three research questions, and conclude with a discussion in Section~\ref{sec:conclusion}.

\section{Existing Methods in Maritime Traffic Surveillance}
\label{sec:related_work}
Intentional shutdown of the AIS transponder to hide illegal fishing or illicit cargo transfer is a common illegal activity in maritime domain, and several methods have been proposed to detect intentional AIS shutdown.
We review these methods in Section~\ref{sec:ShutdownDetection}.
Then, in Section~\ref{sec:MLmethods}, we present the methods processing satellite AIS data using machine learning. 

\begin{figure*}[ht!]
    \centering
        \includegraphics[width=\textwidth]{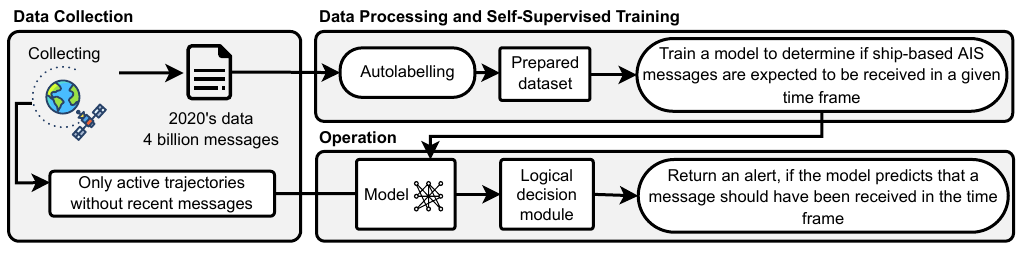}
    \caption{Overview and data flow of our self-supervised learning method for the detection of abnormal missing AIS reception.}
    \label{fig:overview}
\end{figure*}

\subsection{AIS Shutdown Detection}
\label{sec:ShutdownDetection}
In \cite{Mazzarella2017, mazzarella-ais-2016}, Mazzarella \etal design a method to compute the probability of receiving an AIS message depending on the distance to the base station. For that, the signal strength to the base station is used to create a probability map. Also, in \cite{shahir_mining_2019},  Shahir \etal draw a probability distributions for suspected dark fishing.
In \cite{kontopoulos_real-time_2020}, Kontopoulos \etal identify AIS switch-off in near real-time by analysing large streams of AIS messages received from terrestrial base stations. The method relies on the high-quality of the data coming from these sources by predicting precisely when AIS messages are expected for a given ship trajectory in a given cell from a coverage network.
In \cite{singh_machine_2020}, Singh \etal propose to detect AIS switch-off in a labelled dataset extracted from a single base station using a multi-class artificial neural network, that can additionally determine if anomalies are due to power outage. The dataset is composed of \num{132352} messages, corresponding to \num{133} distinct vessels.
These three solutions have proved to achieve excellent results when using high-quality AIS datasets, but they are limited to only detecting intentional AIS shutdown of vessels located in the neighbourhood of base stations. They have not been experimented yet when using satellite-based AIS datasets, which is the only source of AIS available in the open sea. Besides, by using only ground-based AIS message streams, they cannot detect vessels having their AIS already switched off which penetrate into a surveillance zone. Therefore, it seems complementary to combine these methods with solutions adapted for the open sea able to process satellite data. In this context, AIS missing reception is influenced by the spatial and temporal coverage of the satellites, as well as potential signal collisions and weather conditions. These elements make satellite-based datasets very fragmented, with irregular missing reception from a few minutes to several hours. Moreover, when considering trajectories located in the neighbourhood of base stations, the signal coverage of satellites is very poor as compared to the signal from beacons, which makes any comparison or generalisation very difficult. 
In \cite{ford_detecting_2018}, Ford \etal propose a Generalised Additive Model (GAM), adapted to the open sea, which captures space-time variations in AIS gaps during transmissions. The model aims at detecting abnormal AIS missing reception and the overall approach shows good results for an area located between the Australian and Indonesian EEZ border. Interestingly, the approach uses the frequency of transmissions of other vessels to compute the probability of (not) receiving a message. It renders the method more accurate when considering highly dense traffic zones. However, the confidence level of the method starts at a very low level and increases over time because it is based on a probability map. Hence, the solution is well-suited for post-mortem analysis, i.e., analysis of the traffic several hours after it actually happened. In \cite{weimerskirch-ocean-2020}, Weimerskirch \etal use GPS data provided by GPS transponder attached on Albatross birds to track fishing vessels. This solution is original and works in real-time, but it is limited to specific areas where albatross birds live and follow fishing vessels.
In \cite{dafflisio_detecting_2018}, d’Afflisio \etal propose a procedure based on the
Ornstein–Uhlenbeck mean-reverting stochastic process to detect anomalous deviations from
standard maritime routes hidden behind AIS Shutdown. This solution cannot be adapted to fishing vessels that have unpredictable trajectories. It also requires to wait for the re-connection to determine the abnormality. 
In \cite{park-illuminating-2020}, Park \etal make the correlation between the vessel lights at night captured with satellite images and the reported AIS messages received from these ships. Using this correlation, the authors discovered a dark fleet fishing illegally in North Korean waters,  estimated to be worth \$440 million. However, the generalisation of this solution is unfortunately impossible as the vessel lights are usually turned off during the day. While the two last presented solutions are original and useful for evaluating the negative impact of illegal fishing activities, they cannot easily be generalised and cannot offer a robust solution to the problem of automatic intentional AIS shutdown detection. Using hidden information channels for detecting illegal activities is appealing, however, the only source of reliable information available in open waters is satellite-captured AIS data.

In summary, existing analysis methods for detecting AIS switch-off are mostly based on high-quality data coming from base stations, or on probability maps which are relevant only for post-mortem analysis, or else on hidden channels that cannot be generalised. In contrast, our work focuses on near real-time prediction in the open sea based on satellite data that are highly irregular and impacted by the position of the satellites, the number of vessel in the area, the weather and more. In that respect, it is complementary to other methods. Also, it is worth noticing that shared labelled datasets for detecting intentional AIS shutdowns do not exist, which makes fair comparison between different methods very difficult.

\subsection{Machine Learning Applied to AIS}
\label{sec:MLmethods}

The major challenge with AIS messages received by a fleet of surveillance satellites is that they contain irregular patterns and missing data. 
Also, AIS data captured by satellites is noisy because there can be errors in the information entered by vessel captains, or AIS transponders can be of low quality or damaged. 
Previous work that  processes AIS data automatically has focused on detecting anomalies from predicted ship trajectories, trajectory reconstruction, collision avoidance and future traffic evaluation.

Among existing techniques, the usage of probabilistic models of ships' behaviour from historical AIS data has been proposed. For instance, the exploitation of Gaussian mixture models (GMMs)~\cite{dalsnes_neighbor_2018}, grid-based methods~\cite{ristic_detecting_2014}, Markov models~\cite{Fridman2019} or hierarchical neural networks~\cite{Kim2018} have led to significant advances in terms of trajectory reconstruction, trajectory prediction and anomaly detection.
Other work has also focused on enriching vessel information from radar and visual tracking~\cite{bloisi_enhancing_2017}.
For detecting anomalies in AIS message transmission, reconstructing the trajectory of vessels just before the loss of the signal is crucial.
Arguedas \etal use graph-based methods to create a promising lightweight representation of vessel trajectories~\cite{fernandez_arguedas_maritime_2018}. 
Nguyen \etal in \cite{Nguyen2018, Nguyen2019} use a representation that regularises the frequency of messages by completing a dataset with artificially generated missing messages. Interestingly, this approach  deals with noisy data and allows datasets to be used for training models for multitasking learning. However, the datasets do not come from satellites but from beacons at sea. As a limitation, such an artificially completed datasets cannot be used to detect intentional AIS shutdown because the missing messages have been re-introduced for the purpose of regularisation \cite{Nguyen2018,Nguyen2019}.

Other approaches include using ship-tracking probability maps, such as those created by Skauen~\cite{skauen-ship-2019}, to estimate the probability of receiving (or not receiving) a satellite-based AIS message in a given time frame. However, these maps estimate the probability of re-detecting an already detected ship, which implies that one needs to wait a significant amount of time before any action is triggered.

In contrast to these works, we propose to train a self-supervised model for a given vessel, in order to build a representation that  considers the previous messages and the gap between them, as explained in Section~\ref{sec:dataset_creation}. This allows us to propose a method that reveals suspicious cases of intentional AIS shutdowns as soon as a vessel's signal disappears.

\section{Method Overview}
\label{sec:overview}

\begin{figure}[t]
    \centering
    \includegraphics[width=\columnwidth]{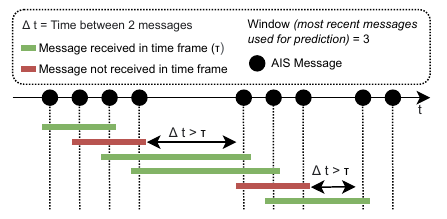}
    \caption{Auto-labelling of trajectories: Green flags indicate trajectories expecting an AIS message in the next time frame $\tau$, while red flags indicate trajectories without an AIS message.}
    \label{fig:reception}
\end{figure}

Before giving an in-depth presentation of our self-supervised method, we introduce a high-level overview of its different components and their connection, as shown in Fig.~\ref{fig:overview}. 
The diagram highlights three distinct processes, namely, \textit{Data Collection}, \textit{Data Processing and Self-Supervised Training} and \textit{Operation}, and their workflow.
First, during \textit{Data Collection}, the satellite operator collects and stores AIS data from geo-marine observation satellites.
In the context of this work, the Norwegian Coastal Administration has provided us access to the AIS data collected by the Norwegian state operator Statsat AS during 2020, which corresponds to approximately 4 billion messages.
As part of the \textit{Data Processing and Self-Supervised Training} process, the collected data is first prepared and processed into a dataset, including the enrichment with self-supervision labelling information. 
As said above, the main issue is to distinguish ordinary vs. abnormal missing AIS reception. For this purpose, we have used the following artefact for implementing our self-supervised training method with the collected data. 
For a given ship in the open sea, given a window of $w$ successive AIS messages (typically $25$ messages, but other window sizes can be considered) and a time frame $\tau$ (typically, $10$ minutes), one can train a model that predicts if an AIS message is expected to be received in the next time frame $\tau$. 
This model will then be useful to compare its prediction with the actual observation performed in near real-time (i.e., using the same time frame $\tau$) regarding the reception of the expected AIS message.
Fig.\ref{fig:reception} illustrates this principle by considering a window of $w=3$ messages. Red-flagged trajectories indicate that no message is expected within $\tau$ while green-flagged trajectories indicate the opposite.

During \textit{Operation}, by using the trained model with the auto-labelled dataset and a logical decision module that analyses real-time observations streamed from the surveillance satellites, it becomes possible to classify \textit{ordinary} versus \textit{abnormal} missing AIS reception.
Every collected vessel trajectory is again preprocessed into the machine learning model input format, although in this context without the addition of self-supervision information that is only necessary for training purposes.
Based on the time frame $\tau$, the logical decision module then classifies the vessel trajectory as either abnormal, highlighting a risk of intentional AIS shutdown, or as ordinary.
In the first case, a specific alert can be escalated to an operator or a downstream system for further investigation, while in the second case issues no alert.

\section{Data Processing}
\label{sec:data}

\subsection{Automatic Identification System (AIS)}
According to AIS technical specifications \cite{itu_technical_2014}, messages are composed of static and dynamic information.
Static fields include, amongst others, international standard vessel identifiers, i.e., MMSI\footnote{Maritime Mobile Service Identity}, IMO ship identification number, vessel name and call sign, ship dimensions and vessel type. 
These static elements, which are entered manually by the vessel captains into the AIS transponder, are automatically transmitted on a broadcast channel every 6 minutes. Also, AIS transponders send dynamic information every 2 to 10 seconds depending on the vessel's speed, or every 3 minutes if the vessel is at anchor. 
The dynamic information includes navigation status, e.g., ``at anchor'' or ``fishing'', vessel position in terms of latitude (lat) and longitude (lon), vessel speed over ground, its direction relative to the north pole, its true heading relative to the magnetic north pole and the timestamp when the message was sent.
There are currently $27$ different AIS message types. 
Of these message types, we only consider the subset that contains the dynamic information, i.e., message types 1, 2, 3, 18 and 19.
Typical AIS transponders have a range of about $20$ to $40$ km. 
The limitation of this range is due to the curvature of the earth and the height at which the antenna is installed on ships.
For vessels too far away from the coast or other AIS observers, AIS satellites collect and forward the messages.

\subsection{Data Collection}

One challenge with AIS data collection is that the radio access scheme defined in the standard creates only 
\num{2200} available time slots every minute for each of the 2 channels %
and receivers can be easily overwhelmed by large AIS reception footprints \cite{skauen-ship-2019}. 
Due to the growing number of AIS transponders and the expansion of the satellite reception area, message collisions can occur and lead to vessel disappearance \cite{eriksen_maritime_2006}.

Our satellite-based AIS dataset, provided by the Norwegian Coastal Administration and collected by Statsat, consists of \num{4050019441} AIS messages from all over the world, which corresponds to the messages collected for the year 2020 (see Table~\ref{tab:data_overview}).
Technical issues on the satellite during the original data collection can explain the difference in the number of messages per month. However, we did not notice any bias in the data or any effect on our experimental results as a result of this difference.

\begin{table}[t]
    \renewcommand{\arraystretch}{1.1}
    \caption{Overview of Satellite AIS Data Collection}
    \begin{tabularx}{\columnwidth}{lcc||lcc}
        \toprule
        Month & \#Msg & Uniq. vessels & Month & \#Msg & Uniq. vess.\\ 
        \midrule
        January  & 138M & \num{46250} & July      & 504M & \num{62389}\\
        February & 113M & \num{43243} & August    & 500M & \num{61070}\\
        March    & 118M & \num{43682} & September & 434M & \num{59583}\\  
        April    & 506M & \num{58603} & October   & 472M & \num{59911}\\
        May      & 513M & \num{61377} & November  & 211M & \num{49284}\\
        June     & 454M & \num{61451} & December  &  80M & \num{40366}\\ 
        \bottomrule
    \end{tabularx}
    \label{tab:data_overview}
 \end{table}

\subsection{Feature Extraction}
\label{sec:feature_extraction}

From each message of the AIS dataset, we select relevant characteristics, namely position ($lat, lon$), the timestamp ($t$), and speed ($s$).
Also, we enrich the characteristics with $\Delta t$ the time difference compared to the preceding message from the same ship, $\Delta D_V$ the difference in meters on the latitude with the preceding message from the same ship, $\Delta D_H$ the difference in meters on the longitude, $D_P$ the distance to the port, $T_D$ the second of the day (between 0 and \num{86399}). We have chosen to split the relative distance with the previous message between $\Delta D_H$ and $\Delta D_V$ to keep the direction of movement.

It is worth noticing that:
\begin{itemize}
     \item \noindent The distance $\Delta D_V$, $\Delta D_H$, $D_P$ are computed with the \textit{Haversine formula}\footnote{The distance of the arc on a sphere is not as precise as the Vincenty formulas in geodesy since the earth is not a perfect sphere, but its computation has the advantage of being vectorisable, which is necessary in the case where the dataset contains a very large number of trajectories.}.
     \item \noindent $\Delta t$ and $T_D$ allow having a better precision on the temporal dimension than that given by the timestamp $t$,
     \item \noindent $\Delta D_V$ and $\Delta D_H$ improve understanding of position over small distances,
     \item \noindent $D_P$ allows filtering the relevant samples (more information on this point in Section~\ref{sec:dataset_creation}).
\end{itemize}
Thus, the vector of characteristics corresponding to a message takes the following form:
$$m = \left[t, lat, lon, s, \Delta t, \Delta D_V, \Delta D_H, D_P, T_D\right]$$

A \textit{trajectory} $\mathcal{E}$ is a temporal sequence of $\mathcal{T}$ successive messages from a ship $\mathcal{E} = [m_1, m_2, \dots, m_T]$. 
It corresponds to the trajectory of a ship's motion in a time window.
Also, note that there is no constraint on the maximum values of $\Delta t$ and $\Delta D$, between two successive messages within a trajectory.

\subsection{Dataset Creation}
\label{sec:dataset_creation}

\begin{figure*}[!t]
    \includegraphics[width=\textwidth]{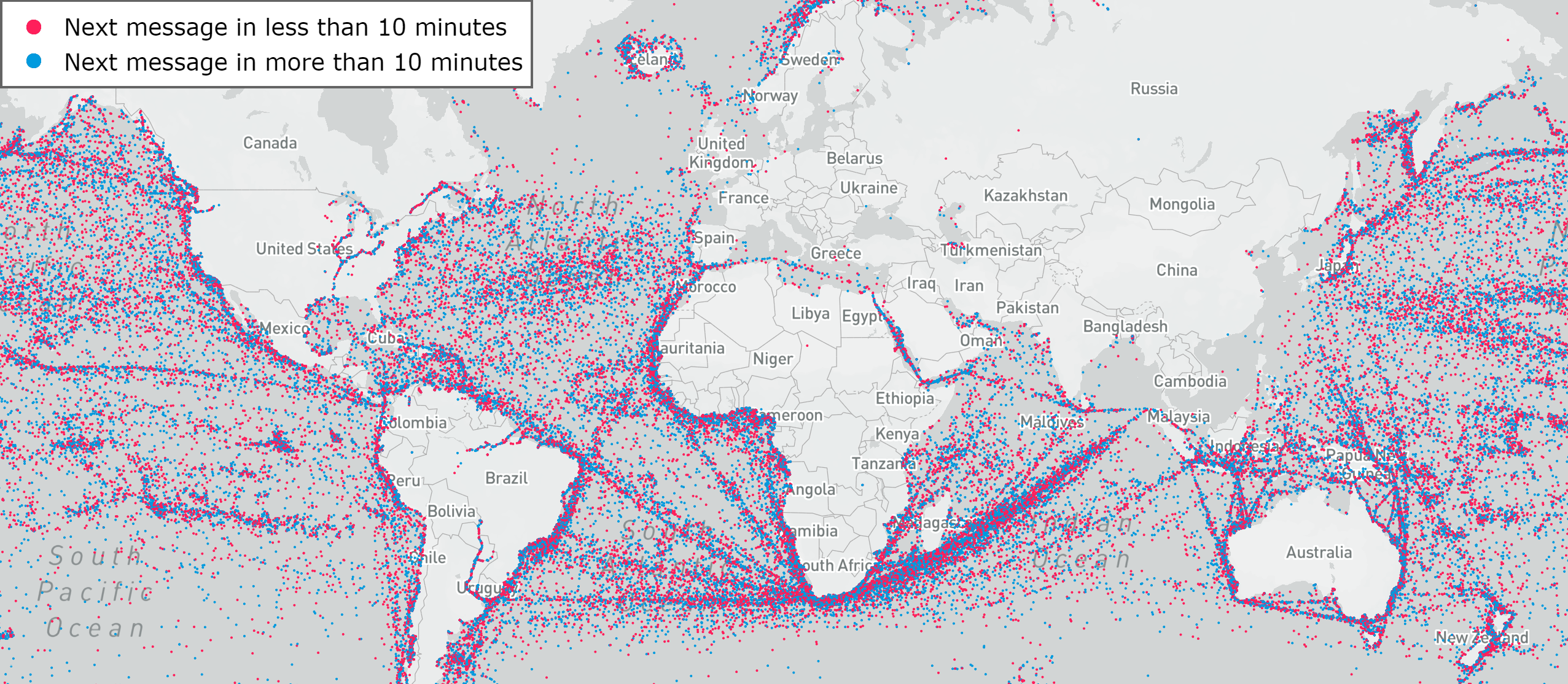}
    \caption{Distribution of January's samples over the surface of the globe. Each point shows the location of the most recent message and whether the next message will be received within 10 minutes.}
    \label{fig:repartition}
\end{figure*}

A total of \num{500000} trajectories has been extracted for each month, creating 12 datasets, equally balanced between two classes: 1. trajectories including a message received within the time frame; 2. trajectories without any message received within the time frame Fig.~\ref{fig:repartition} shows the distribution of the samples on the surface of the globe for January 2020.
For the selection of trajectories in the dataset, we set two conditions:
\begin{enumerate}
    \item The AIS vessel trajectory history must consist of at least $50$ messages to exclude vessels not sufficiently represented over the entire year of the dataset. This corresponds to about a 1-minute history for a moving vessel. Indeed, these trajectories are too short for any relevant generalisation by the model; 
    \item The ship's loss of reception must be at more than $5$ kilometres from a port ($D_P>5km$). This eliminates examples of intentional (legal) AIS shutdown that take place in ports. It is also relatively easy for the coastal administration to control vessels without AIS transmission in ports. To determine the distance to the nearest port, we calculate the distance of the arc between two points on a sphere with a database of nearly \num{30000} ports compiled by the organisation \textit{Global Fishing Watch}\footnote{\url{https://globalfishingwatch.org/datasets-and-code/anchorages/}}.
\end{enumerate}

\begin{table}[t]
    \renewcommand{\arraystretch}{1.1}
    \centering
    \caption{Statistics for Jan. 2020 dataset (500k trajectories) show a wide variety of trajectory characteristics. ($\Delta d$~=~Dist. between messages , $\Delta t$~=~Time between messages, $\Sigma d$~=~Trajectory length and $\Sigma t$~=~Trajectory duration)}
    \begin{tabularx}{\columnwidth}{lrrrrrrr}
        \toprule
        Percentile & 1st & 5th & 25th & 50th & 75th & 95th & 100th\\ \midrule
        $\Delta d$ [m] & 0 & 0 & 0 & 48 & 307 & 5.6k & 19.7M\\
        $\Delta t$ [s]   & 1 & 4 & 11 & 52 & 372 & 4.0k & 2.1M\\
        $\Sigma d$ [km]         & 0 & 0 & 5.8 & 40.4 & 118.5 & 555.5 & 648.1\\  
        $\Sigma t$ [h]        & 0.42 & 1.28 & 4.1 & 8.58 & 22.89 & 147.44 & 722.22 \\
        \bottomrule
    \end{tabularx}
    \label{tab:dataset_stat}
\end{table}

The datasets used for training, validation, and testing of the model are separated by date, e.g., train on the January data with 10\% of the February data for validation, and then test on the rest of the February data. 
Another alternative is to separate by ships, i.e., train on one group of vessels and test on another, but the temporal separation has the advantage of being closer to the operational use of the model, since the objective is to make predictions based on past data to predict upcoming missing AIS reception.
A further alternative would be to split by geographical regions, but this would either require a strong regional generalization or result in a highly region-specific model, whereas we aim for a globally applicable model.
Therefore, the model will be trained according to the described separation by date.

Table~\ref{tab:dataset_stat} shows the range of duration and distance between two consecutive messages along with the total duration and distance trajectories.
We observe the dataset to be heterogeneous with strongly varying time spans and distances between messages as well as drastically different trajectories, explained both by different vessel types and irregularities in AIS reception and therefore potentially long gaps between messages.
Noticeably, 1\% of the messages are sent with a time-interval of 1s or less, which is abnormal w.r.t. AIS protocol. In fact, our data is noisy and the noise is mostly due to inappropriate reuse of MMSI numbers. Skauen {\it et al.} estimate than in between 0.5\% to 2\% of the MMSI numbers are reused \cite{skauen_quantifying_2016, skauen-ship-2019}.

\section{Trajectory Preprocessing and Self-Supervised Machine Learning Model}
\label{sec:model}

Fig.~\ref{fig:archi} shows an overview of the architecture of the deep learning model. 
Firstly, the preprocessing step encodes and normalizes the input trajectory $\mathcal {E}$ to simultaneously handle long distances between two messages and to maintain sufficient precision for those messages having only few seconds of difference. 
It also allows us to dissociate the trajectory from the absolute time and position toward a generic representation.
Secondly, the two output vectors of the preprocessing step are transferred to the deep learning model that is trained to classify whether a message should be received or not.

\subsection{Trajectory Preprocessing}

The preprocessing block in the Fig.~\ref{fig:archi} is essential to our approach.
Following the feature extraction describe in Section~\ref{sec:feature_extraction}, the input trajectory $\mathcal{E}$ is divided into two vectors, $\mathcal{V_H}$ the history of AIS messages which contains the information relating to the previous message and $\mathcal{V_L}$ the most recent position. 
Then $\mathcal{V_H}$ and $\mathcal{V_L}$ go through a normalization layer $\mathcal{N}$. 
The input is divided to have an encoding that considers the possible small distances that can be found between two messages while maintaining very high precision on the starting position of missing AIS reception. 

\subsubsection{Message History}

To represent the message history $\mathcal {V_H}$ generically, the normalization layer $\mathcal{N}$ first removes the absolute position in latitude and longitude and the absolute time represented by the timestamp. 
Instead, the representation only considers the time and distance differences relative to the previous message.
Since the model is supposed to work in relation the most recent received message of a vessel's trajectory we can therefore create a relative representation space and work time-independently.
In addition, the second of the day $\mathcal{S_D}$ is added to strengthen the detection of temporal patterns. 
A cyclic normalisation $\mathcal{N_C}$ (Eq. \ref{nc}) is applied to cyclic fields such as the $\mathcal{S_D}$, a linear normalisation $\mathcal {N_L}$ (Eq. \ref{nl}) is applied to limit values such as speed $\mathcal{V}$. 
For $\Delta t$, $\Delta D_V$ and $\Delta D_H$, a logarithm is applied (Eq. \ref{vp}).
 
\begin{equation}
    \mathcal{N_C} = \left[\sin{ \left(\frac{2 \pi (x - \min x)}{\max x - \min x}\right)}, \cos{\left(\frac{2 \pi (x - \min x)}{\max x - \min x}\right)}\right]
    \label{nc}
\end{equation}
\begin{equation}
    \mathcal{N_L} =\frac{(x - \min x)}{\max x - \min x}
    \label{nl}
\end{equation}
\begin{equation}
\begin{aligned}
\mathcal{N}(\mathcal{V_H}) = [log(\Delta t), log(\Delta D_V), log(\Delta D_H), \\
\mathcal{N_C}(\mathcal{S_D}), \mathcal{N_L}(\mathcal{V})]
\label{vp}
\end{aligned}
\end{equation}
\begin{equation}
\begin{aligned}
\mathcal{N}(\mathcal{V_P}) = [ \mathcal{N_L}(lat_{Deg}), \mathcal{N_C}(lat_{Min}), \mathcal{N_C}(lat_{Sec}), \\
\mathcal{N_C}(lon_{Deg}), \mathcal{N_C}(lon_{Min}), \mathcal{N_C}(lon_{Sec}), \dots ]
\label{vn}
\end{aligned}
\end{equation}

\subsubsection{Most Recent Position}
The vector $\mathcal{N}(\mathcal{V_P})$ consists of 11 values allowing the model to have maximum precision on the position of the vessel at the most recent message received. 
$\mathcal{N}$ decomposes latitude and longitude into Degree-Minute-Second and then normalises them cyclically to maintain continuity (Eq. \ref{vn}), for example, when a ship passes longitude $180$ at $-180$.

\subsection{Model Architecture}

Our method to detect the abnormal missing AIS reception is based on \textit{transformer} models, a deep learning architecture.
Transformer models have demonstrated compelling results in the field of Natural Language Processing (NLP) \cite{Devlin2019} and the field of computer vision \cite{Dosovitskiy2020}.
The suitability of the transformer architecture for dealing with sequential data makes it a good candidate for the analysis of time series. In the context of AIS, the message history can be seen as a time series \cite{Li2019, Lim2021}.
The transformer architecture is a variant of \textit{self-attention networks} \cite{bahdanau-neural-2015} for processing sequential data, however it does not involve a recurrent network architecture \cite{hochreiter-long-1997}. 
This is possible due to the \textit{multi-head self-attention} \cite{Vaswani2017} mechanism, that allows it to attend and weight multiple parts of the sequence differently in parallel. Thereby, it allows to model both long-term and short-term dependencies within the input sequence \cite{cheng-long-2016}. 

We specifically use the encoder part of the transformer model to extract a general fixed-size \textit{representation} vector, that can be used in a generic way for different applications in maritime surveillance, here for the detection of abnormal missing AIS reception.
More specifically, the model consists of two transformer blocks, following on the model of \cite{Vaswani2017}, to encode a variable-length input sequence $\mathcal{V_H}$ together with a positional encoding into the first part of the representation vector $\mathcal{R}$.
Choosing \num{64} as the size of the representation vector R results from a tradeoff between the model complexity and an attempt to prevent the classical overfitting risk and has been determined through preliminary experiments.

This first part is concatenated with the second input, the most recent position $\mathcal{V_L}$, into the overall fixed-size representation vector.
This vector is then processed by the final, task-specific subnetwork.
Here, this subnetwork consists of three dense layers of respective size (100, 50, 50)  with ReLu activation and 10\% dropout and a final layer to classify whether a message should be received within the given time frame.

\begin{figure}[t]
    \centering
    \includegraphics[width=0.95\columnwidth]{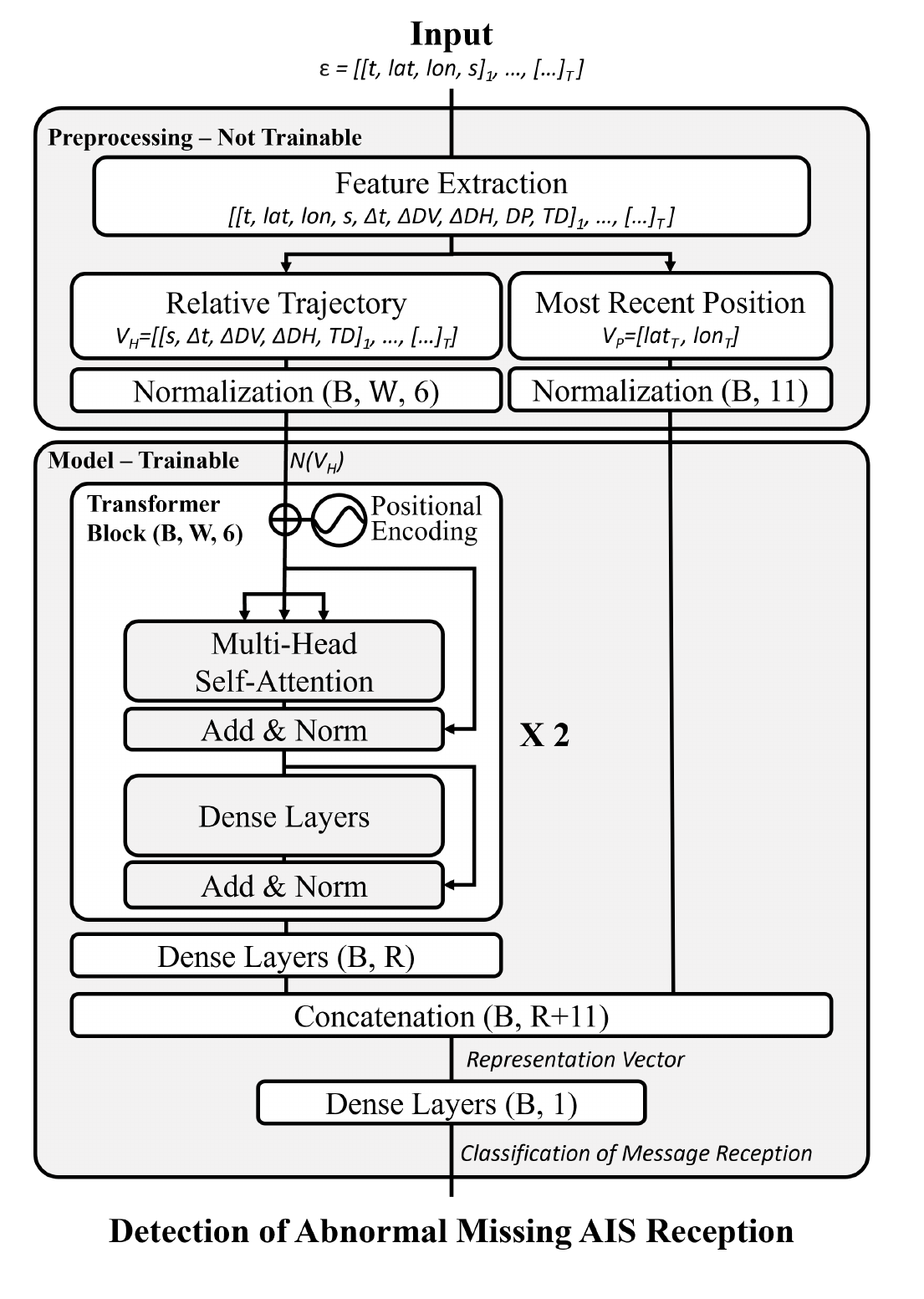}
    \caption{Model architecture: The relative vessel trajectory is encoded via a transformer network and forms together with the most recent position a representation vector for classification.}
    \label{fig:archi}
\end{figure}

\subsection{Detection of Abnormal Missing AIS Message Reception}
Given a ship trajectory and a time frame $\tau$, the model is trained to predict whether an AIS message from the vessel shall be received or not, within $\tau$. 
Using this model, it becomes possible to classify ship trajectories as ordinary or abnormal in near real-time. More precisely, there are four distinct situations for a given trajectory:
\begin{enumerate}
    \item If the model predicts that no AIS message is expected and no message is actually received during the time frame $\tau$, then the trajectory can safely be classified as \textit{ordinary}. The model has correctly learnt that the ship has an ordinary AIS disconnection due to any acceptable cause (i.e., reception conflict, loss of satellite coverage, etc.);  
    \item If the model predicts no AIS message and a message is actually received within $\tau$, then there is a prediction error in the model. The trajectory can be safely classified as \textit{model error};
    \item If the model predicts that an AIS message shall be received within $\tau$ and a message is actually received, then the trajectory can safely be classified as \textit{ordinary};
    \item If the model predicts that an AIS message is expected within $\tau$ and no message is actually received, then the trajectory can be classified as \textit{abnormal}. These latter cases are the most interesting for the coastal administration, whom can then proceed to a detailed analysis of the ship trajectory in order to confirm or denial the abnormal missing AIS reception. In this case, the complete vessel AIS history is reported.
\end{enumerate}
During operation, only the missing AIS reception that exceed the time frame are analysed. This is corresponding to the cases 1) and 4)
Note that having a prediction model with high precision is crucial here to avoid the production of too many false negatives or false positives. The detailed examination of a ship trajectory takes time and there is a challenge to report only suspicious trajectories which are likely to correspond to actual intentional AIS shutdown.  
Note also that the time frame must be exhausted in order to classify ship trajectory. Besides justifying the terminology ``near real-time'', it also justifies the selection of a value for $\tau$ which is compatible with current practices of the coastal administration. Indeed, it will be useless to classify ship trajectories as suspicious after a long time like $24$ hours. 

\section{Experimental Evaluation}
\label{sec:results}

In this section, we evaluate the capabilities and robustness of the proposed model. For that, we examine different configurations of the trained models, such as their architecture and the training set size, the selection of the most appropriate time-horizon for improving the detection of missing AIS messages. More precisely, we investigate these three research questions:

\begin{enumerate}
    \item[\textbf{RQ1}] Is the method effective enough to detect abnormal AIS shutdown?
    \item[\textbf{RQ2}] How stable is the model over time, and is it affected by dataset shift?
    \item[\textbf{RQ3}] How robust is the model and its effectiveness to configuration changes, i.e., changes in training set size, message window size, model architecture, or predicted time horizon?
\end{enumerate}

\subsection{Experimental Setup}
\label{sec:experimental_setup}
In our experiments, the batch size $\mathcal{B}$ is equal to $128$, the message window size $\mathcal{W}$ is $25$ and the representation vector size $\mathcal{R}$ is $64$ (see Fig.~\ref{fig:archi}).
With these hyperparameter values, the model $\mathcal {M}$ has a total of \num{470821} trainable parameters.
For model training, the dataset contains \num{200000} trajectories with 10 minutes time horizon labels.
The model is trained over a maximum of $200$ epochs with each $1562$ batches with early stopping once the validation loss converged.
All training is performed with the configuration described above, unless stated otherwise. Section~\ref{sec:model_robustness} compares the results for the most impactful hyperparameters.
All our experiments are run on an NVIDIA DGX-2 with an Intel Xeon Platinum 8168 CPU with 2.7 GHz and 24 cores, using one NVIDIA Tesla V100 graphics card\footnote{Provided by the eX3 research infrastructure: \url{https://www.ex3.simula.no/}}.
One epoch of model training takes approximately $17$s.
We have evaluated the model stability by running each model prediction five times with different initial random models. As said in Section~\ref{sec:dataset_creation}, the one-year dataset is separated into $12$ time-windows, one per month. 
 
\textbf{Evaluation Metrics.} Three metrics are used to compare the models: 1) the \textit{accuracy} (Eq.~\ref{acc}), which measures the overall capability of the model to predict whether a message should be received or not; 2) the {\it positive predictive value} (Eq.~\ref{ppv}), or \textit{precision}, focuses on the model capability to predict whether it is normal to not receive a message or not; and 3) the negative predictive value (Eq.~\ref{npv}), which focuses on the error made by the model when it predicts that we should not receive a message, but we receive one.

Formally speaking,
\begin{equation}
    \text{Accuracy} =\frac{TP + TN}{TP+FP+TN+FN}
    \label{acc}
\end{equation}
\begin{equation}
    \text{Positive predictive value (PPV)} = \frac{TP}{TP+FP}
    \label{ppv}
\end{equation}
\begin{equation}
   \text{Negative\ predictive\ value (NPV)} = \frac{TN}{TN+FN}
    \label{npv}
\end{equation}
where TP stands for {\it True Positive}, TN for {\it  True Negative}, FP for {\it False Positive} and TN for {\it False Negative}.

\subsection{RQ1: Model Effectiveness}
\label{sec:model_effectiveness}

There are no datasets for evaluating our model precisely on the task of detecting voluntary AIS shutdown. Therefore, we have reversed the question into predicting whether a message is received within a time frame or not in order to train a model in a self-supervised manner. 

Therefore, it is important to note that the model accuracy corresponds to the prediction of message reception within a time frame and not the prediction of voluntary AIS shutdown. 
For this particular task, the combination of the model and the encoder presented in this paper has shown the high accuracy between 99.5\% and 99.8\%. Table~\ref{tab:confusion} shows the confusion matrix for prediction results of \num{4500000} trajectories. %
\textbf{False negative (F. neg.)} means that the model predicted a non-reception of a message, while the message is received; \textbf{False positive (F. pos.)} means that the model predicted a reception of a message, while the message is not received.

\begin{table}[t]
    \renewcommand{\arraystretch}{1.1}
    \centering
    \caption{RQ1: Confusion matrix (in \%), on 4.5M trajectories of vessels disappearing more than 10 minutes.}
    \label{tab:confusion}
    \begin{tabular}{clrrr}
        \toprule
        & & \multicolumn{2}{c}{Annotation (Ground Truth)} & \\
        \cmidrule{3-4} & & Yes & No & \\ \midrule
        \parbox[t]{2mm}{\multirow{2}{*}{\rotatebox[origin=c]{90}{Pred.}}} & Yes & \cellcolor{gray!25}49.96 & (F. pos.) 0.20 & PPV = 99.60 \\
        & No & (F. neg.) 0.04 & \cellcolor{gray!25}49.80 & NPV = 99.92\\        \bottomrule
    \end{tabular}
 \end{table}
 
 \begin{table*}[t]
    \renewcommand{\arraystretch}{1.1}
    \caption{RQ2: Stability of the model (in \%). Small seasonal effects throughout the year, but the model is generally stable.}
    \label{tab:model_stability}
    \begin{tabularx}{\textwidth}{lCCCCCCCCCCC|c}
    \toprule
        Month & Feb & Mar & Apr & May & Jun & Jul & Aug & Sep & Oct & Nov & Dec & Average\\
        \midrule
        Model trained on previous month  & 99.61 & 99.87 & 99.69 & 99.70 & 99.78 & 99.72 & 99.75 & 99.74 & 99.84 & 99.63 & 99.90 & 99.75 \\
        Model trained once on January   & 99.61 & 99.81 & 99.63 & 99.61 & 99.63 & 99.63 & 99.61 & 99.59 & 99.64 & 99.71 & 99.90 & 99.67 \\
    \bottomrule
    \end{tabularx}
 \end{table*}

The 0.04\% of detected false negatives can only be caused by a model error. On the other hand, the 0.2\% false positives can either be due to a model error or an anomaly. 
In our results, \num{8868} of trajectories are returned as anomalies out of \num{2250000} trajectories that are annotated as missing AIS messages. However, there is no precise way to evaluate and filter the proportion of accurate missing AIS receptions among the \num{8868} of total model predictions for a full year covering the whole globe. We discuss model errors in Sec.\ref{sec:discussion}.

\subsection{RQ2: Model Stability over Time \& Dataset Shift}
\label{sec:model_stability}

The second research question addresses the model stability over time and possible effects of dataset shift \cite{Quinonero-Candela2008}.
A trained model has potentially the highest accuracy for trajectories that are close in time to the period covered by the training data, since it is expected to be most similar to the data within the training set, i.e., it is from a similar distribution.
However, over time this distribution might change, e.g., due to seasonal effects.
To investigate the general stability of the model over time and a potential negative impact from dataset shift, we perform two evaluations:
\begin{enumerate}
     \item We train the model on one month and evaluate it on the next month, e.g., training on January and evaluation on February, repeated for all months of the year;
     \item We train the model once on the January dataset and evaluate its accuracy on the other months of the year.
\end{enumerate}

The results are shown in Table~\ref{tab:model_stability}.
We observe a small dataset shift, as indicated by the slightly lower accuracy of the fixed model trained on the January data.
For each individual month this difference in accuracy is not more than 0.15\% and in the average over the year not more than 0.08\%.
The variation in accuracy is accounted for by seasonal effects.
The largest accuracy differences occur in the summer months, i.e., June to September, whereas towards the end of the year the accuracy gap is closer or even non-existent in December.

We conclude from this experiment that the model is generally stable and that there is no need for frequent re-training.
Since the encoding of the vessel trajectory works on relative time and position information, there is no strong dependency or link to the absolute time of the training data, but just the vessel traffic patterns, which are subject to seasonal adjustments.
However, it is advisable to consider a training dataset that spans multiple months or seasons of data to further improve the model stability.

\subsection{RQ3: Model Robustness to Configuration Changes} 
\label{sec:model_robustness}

As previously introduced, our model allows to be adjusted through a number of configuration decisions and parameters.
Those include both parameters that affect the operation in maritime surveillance, i.e., the horizon of time during which an AIS message is expected, and technical decisions, i.e., the dataset size used for training or the model architecture.
To analyse the robustness of the model, we vary each of these parameters individually to identify the sensibility of the overall model to this parameter.

\subsubsection{Dataset Size}
\label{sec:experiment_dataset_size}

As the first parameter, we vary the dataset size used for training from 10k to 500k trajectories of each 25 messages.
As shown in Table~\ref{tab:acc_dataset_size}, the training benefits from more data, but the effect diminishes after \num{100000}, while the training time continues to increase approximately linearly.
We therefore select a dataset size for our experiments of \num{200000} to balance accuracy and training cost.

\begin{table}[t]
    \renewcommand{\arraystretch}{1.1}
    \caption{RQ3: Accuracy by dataset size}
    \label{tab:acc_dataset_size}
    \begin{tabularx}{\columnwidth}{lCCCCCC}
    \toprule
        Dataset Size & 10k & 50k & 100k & 150k & 200k & 500k\\
        \midrule
        Accuracy (\%) & 95.35 & 98.79 & 99.32 & 99.24 & 99.38 & 99.69\\
    \bottomrule
    \end{tabularx}
\end{table}
 
\subsubsection{Window Size}

\begin{figure}[t]
    \includegraphics[width=\columnwidth]{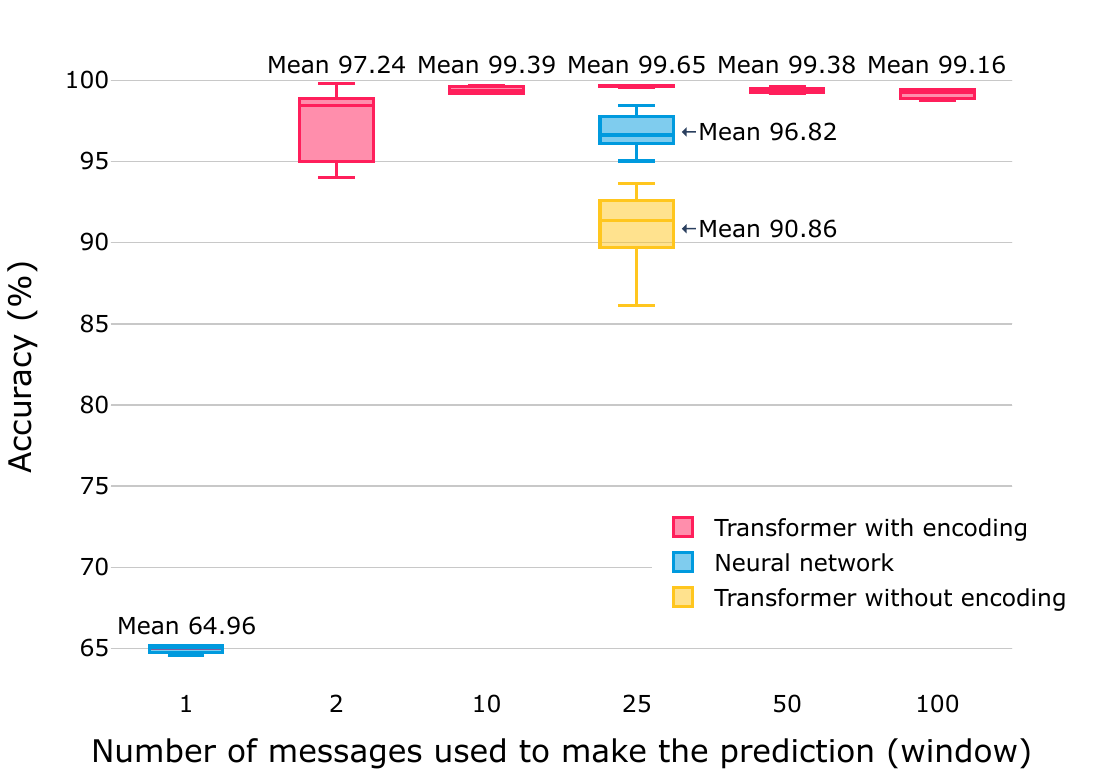}
    \caption{RQ3: Accuracy for different model architectures and message window sizes.}
    \label{fig:Accuracy}
\end{figure}

The window size controls how much of a vessel's history the model receives as input.
Fig.~\ref{fig:Accuracy} shows that some history is relevant to identify vessel behaviour, but that the effect saturates and a higher window size only increases model complexity but reduces the accuracy.
For our experiments, we selected a window size of 25 messages, which showed the best accuracy.
However, it should be noted that the window size is likely to be specific for the surveillance application, e.g., for other applications such as detecting missing AIS reception, a larger window size can be more beneficial.

\subsubsection{Model Architecture}
\label{sec:experiment_model_evaluation}

In this experiment, we vary the model architecture and alternatively select a standard feed-forward neural network and a transformer model without AIS message encoding.
The feed-forward neural network consists of 5 layers with 10/20/25/20/10 neurons, selected by hyperparameter search. 
Since the input layer needs to be adjusted for each input size, we opt to test it with one input message, as the most naive baseline, and 25 input messages, as selected for our main architecture.

The results are again shown in Fig.~\ref{fig:Accuracy}.
While the naive baseline is not competitive, it highlights the necessity of considering a multi-message window size.
For the direct comparison of all three architectures, the feed-forward neural network achieves an accuracy of 96.82\%, which exceeds the 90.86\% accuracy of the transformer without the custom encoding, whereas the full model achieves 99.65\% accuracy.
This result underlines the importance of the dedicated message encoding in combination with the transformer architecture.

\subsubsection{Horizon of Time}

The horizon of time parameter is closely linked to the model's operational use, as it defines the threshold after which the missing reception of AIS messages is deemed interesting or relevant.
We evaluate five time horizons: 1, 5, 10, 30, and 60 minutes.
The model is both trained and tested with these time horizons.

\begin{figure}[t]
    \includegraphics[width=\columnwidth]{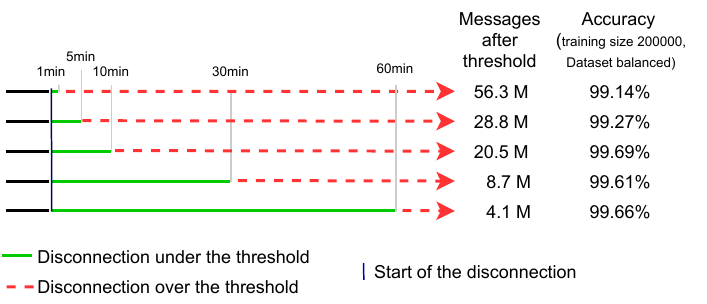}
    \caption{RQ3: Comparison of five different time horizons. Total messages received after each threshold from the whole February data with 138 million messages and accuracy for the model trained with a subset of 200,000 trajectories.}
    \label{fig:Horizon}
\end{figure}

The results are shown in Fig.~\ref{fig:Horizon} with no major difference in accuracy between the different time horizons.
Therefore, it is the choice of the operator between a shorter reaction time with a potentially higher number of misclassifications as anomalies versus a longer reaction time and lower risk of misclassifications.
Even if the accuracy is high, the quantity of missing messages is significant too, and a small prediction error can represent a large number.
If we compare the examples over the threshold value in our dataset, the number of missing messages with a time horizon of 1 minute exceeds those with a time horizon of 60 minutes by a factor of 10.
After discussions with domain experts, we fixed the horizon of time for our experiments to 10 minutes. 
This balances both the number of potential false alerts and the reaction time, which is still short in the context of satellite-based maritime surveillance.

\subsection{Near-Real Time Detection}

Detecting intentional AIS shutdown is highly relevant for maritime surveillance monitoring and any alerting system has to react to potential anomalies in (near) real-time. Technically, by using the hardware components mentioned in Section \ref{sec:experimental_setup}, our implementation processes \num{5400000} samples in \num{409} seconds, i.e., in less than 7 minutes. It corresponds to a processing rate of \num{13191} predictions per second, a significantly higher rate than the rate of messages collected per second by the four satellites used in our experiment. These satellites gather approximately 100 to 150 messages per second, with only a subset of these messages being potential cases of intentional AIS shutdowns.

In practice, real-time detection of intentional AIS shutdown is subject to certain limitations. One of these limitations is the selection of time-horizon. In order to reduce the number of false positives, it is necessary to wait for a user-selected time-horizon, which has been optimally evaluated to 10 minutes in our experiments. 
Another aspect that affects real-time detection, independent of our solution, is the method by which the data is collected. 
The AISSat and NorSat constellations use the Svalbard ground station in Norway to collect data on all 15 of the satellites' polar orbital passes each day. In the worst-case scenario, the data is collected 96 minutes later, causing an inherent delay in processing and detecting intentional AIS shutdown events.

\subsection{Discussion}\label{sec:discussion}
In this study, we achieved excellent accuracy in predicting whether we should receive an AIS message within a specific time frame. The key elements contributing to this high accuracy are the combination of transformer model with the encoding, and the size of the auto-labelled dataset. However, detecting intentional AIS shutdown with high accuracy remains a challenging task, mainly because labelled datasets with ground-truth information do not exist. 
The approach we present in this paper deals with this problem, but we still need to take two limitations into account, which are not unique to the detection of intentional AIS shutdown but common among the setting of limited labelled data in machine learning.
First, to cope with the absence of a labelled dataset, our model is trained in a self-supervised manner with auto-labelled data to predict when the next AIS message shall be received from a given vessel.
If there is a divergence between the prediction and the real-time available information, then an alert is reported. 
Even though we predict next message reception with a very high accuracy, the model does not provide guarantees on the accuracy to detect intentional AIS shutdown. The availability of a labelled dataset would aid in the training of a model with even higher accuracy, although this would incur high cost to create the dataset in the first place.
Second, the effectiveness of our method mirrors the diversity of the training dataset and the capability of the model to generalise from patterns observed during training.
In our context, if an illegal activity is mimicking the most standard vessel behaviour perfectly, it will be highly difficult to detect. 
Similarly, if the behaviour is entirely different from anything seen during training, although this is likely to be detected as an anomaly in a AIS processing pipeline.
In order to investigate the possible materialisation of these issues, we analysed $8868$ trajectories detected as potential anomalies, such as described in Section~\ref{sec:model_effectiveness}. 
By grouping multiple vessels which present the same anomaly, we were able to detect suspicious patterns.
Among these clusters of anomalous trajectories one of them particularly caught our attention. 
It corresponds to a cluster of around 50 fishing vessels shutting down their AIS transponders repeatedly, next to Argentina's Exclusive Economic Zone (in between 10 to 20 times over the year 2020). 
Signal loss can last for multiple days before a vessel reappears again close to the border of the zone.
From the AIS-message based trajectory, it is observable that these vessels go to anchor in Montevideo, Uruguay, before and after going to the Argentina border where they disappear. The alert triggered by our model was later confirmed in a 2021 report by Oceana~\cite{mustain_oceana_2021}.

\section{Conclusion}
\label{sec:conclusion}
Intentional AIS shutdowns are performed to hide illegal activities in open sea, where vessel traffic systems can only rely on satellite AIS.
In this article, we have presented a method for finding suspicious ship trajectories containing intentional AIS shutdown in near-real-time using self-supervised deep learning based on the analysis of satellite AIS data only. 
The model is trained to detect abnormal missing AIS message reception without requiring the provision of a labelled dataset, which would be costly and labor-intensive.
The training process is instead designed to use a self-supervision technique, where the transformer neural network is trained with auto-labelled training data, that is generated from raw AIS data.

Our experimental evaluation shows that the method can predict expected AIS message reception with 99.5\% accuracy on previously unseen test data.
Additional experiments further underline the robustness of the method under different configurations as well as its stability over longer time periods, avoiding the need for continuous re-training.
Finally, we were able to reproduce real-world reports of intentional AIS shutdown using our method.

For future work, we foresee two possible directions to reduce model errors. First, we envision incorporating feedback from the coast guard administration on reported abnormal trajectories to discard false positives and enhance the quality of auto-labeled training data (i.e., by discarding trajectories with ordinary AIS shutdowns which are wrongly reported as suspicious by the model).
Second, our model's performance could be further improved by adding contextual information from both micro- and macro-level data sources. Specifically, this could involve examining the behavior of nearby ships, integrating additional VHF Data Exchange System data, and incorporating optical or infrared satellite imagery for more targeted investigations. Moreover, the inclusion of macro-level factors like weather conditions, sea currents, and vessel density could offer a comprehensive understanding of AIS behavior, thereby improving model robustness.

\printbibliography

\begin{IEEEbiography}[{\includegraphics[width=1in,height=1.25in,clip,keepaspectratio]{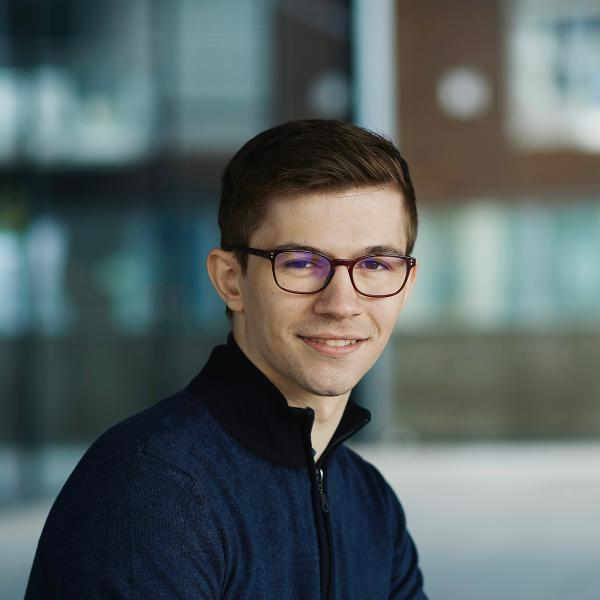}}]{Pierre Bernab\'{e}}{\,} is a PhD student at Simula Research Laboratory in Oslo, Norway. Affiliated to the University Bourgogne Franche Comté (UBFC). His research interests are in the application of machine learning techniques to support maritime surveillance by detecting anomalies in AIS communications. Contact him at pierbernabe@simula.no.%
\end{IEEEbiography}%
\begin{IEEEbiography}[{\includegraphics[width=1in,height=1.25in,clip,keepaspectratio]{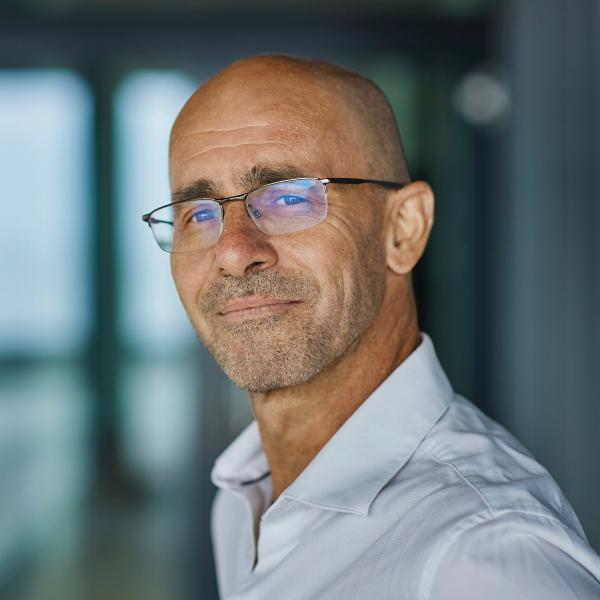}}]{Arnaud Gotlieb}{\,}is Chief Research Scientist at Simula Research Laboratory in Norway. His research interests lie in the application of artificial intelligence to the validation of software-intensive systems, cyber-physical systems including industrial robotics and autonomous systems. Dr. Gotlieb has co-authored more than 120 publications in Artificial Intelligence and Software Engineering and developed several tools for testing critical software systems. Contact him at arnaud@simula.no.%
\end{IEEEbiography}%
\begin{IEEEbiography}[{\includegraphics[width=1in,height=1.25in,clip,keepaspectratio]{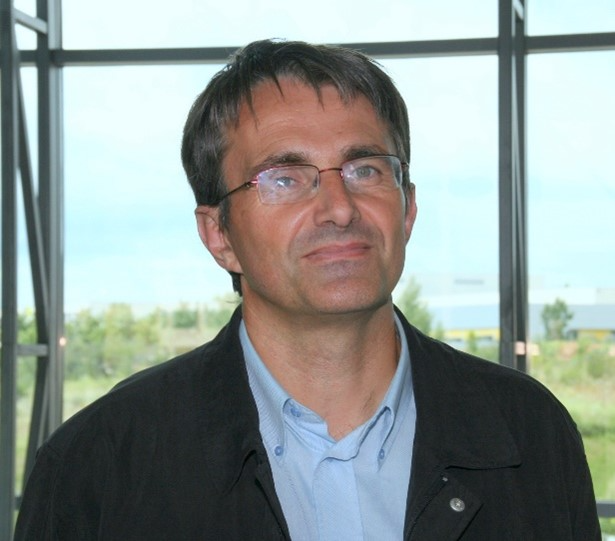}}]{Bruno Legeard}{\,} is Professor at UBFC – Institut FEMTO-ST, Scientific Advisor and co-founder of Smartesting. Bruno has more than 20 years’ expertise in Model-Based Testing/Model-Based Security Testing (MBT/MBST) and its introduction in the industry. His research activities mainly concern features about automation of Model-Based Test case generation using AI techniques. His research results in more than 100 scientific and industrial publications based on MBT and MBST. He is author of three books disseminating Model-Based Testing in the industry. ‘Practical Model-Based Testing’ has more than 1200 citations. Contact him at bruno.legeard@femto-st.fr.%
\end{IEEEbiography}%
\begin{IEEEbiography}[{\includegraphics[width=1in,height=1.25in,clip,keepaspectratio]{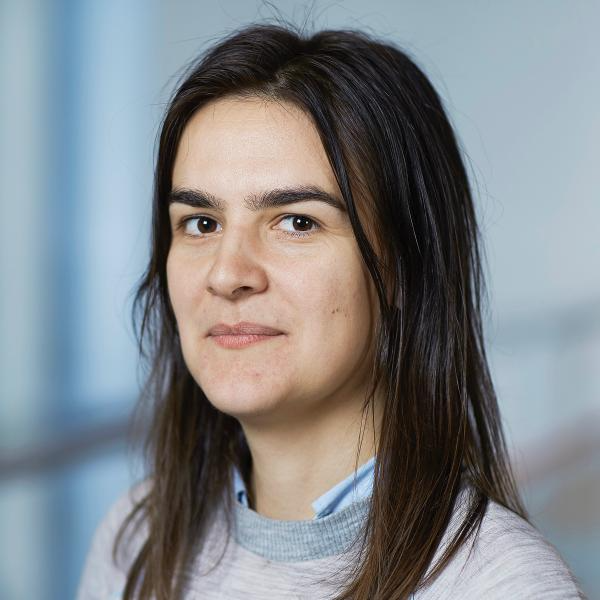}}]{Dusica Marijan}{\,}is a Senior Research Scientist at Simula Research Laboratory in Oslo, Norway. Her research interests are in software engineering, with focus on improving software quality with artificial intelligence techniques. Prior to Simula, she worked as a Senior Software Engineer in the consumer electronics industry. Contact her at dusica@simula.no.%
\end{IEEEbiography}%
\begin{IEEEbiography}[{\includegraphics[width=1in,height=1.25in,clip,keepaspectratio]{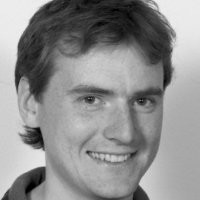}}]{Frank Olaf Sem-Jacobsen}{\,} Frank Olaf Sem-Jacobsen was born in Norway in 1979. He has a PhD in computer science from the University of Oslo, Norway, with focus on fault tolerance in high-performance interconnection networks (both intra datacenter and on-chip). After a period as postdoctoral fellow at Simula Research Laboratory, Frank Olaf joined Space Norway/Statsat as a systems engineer. In this position has built the mission control centre used to control the now five government owned small satellites that gather AIS data around the world 24/7. Contact him at Frank.Sem-Jacobsen@statsat.no.%
\end{IEEEbiography}%
\begin{IEEEbiography}[{\includegraphics[width=1in,height=1.25in,clip,keepaspectratio]{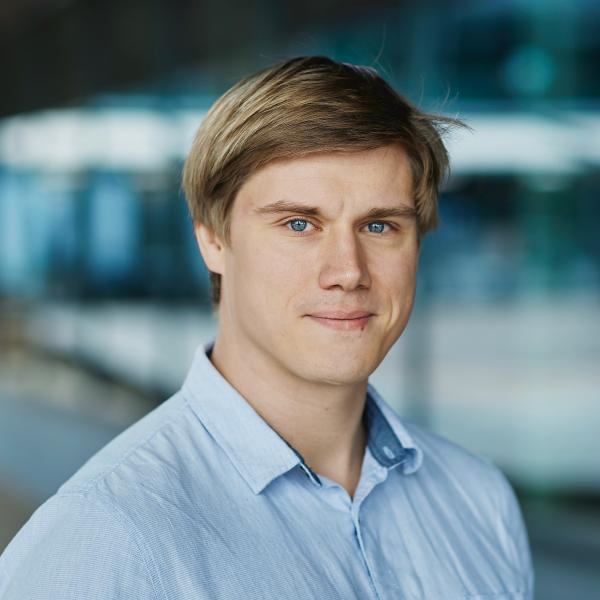}}]{Helge Spieker}{\,}is a Research Scientist at Simula Research Laboratory in Oslo, Norway. His research interests are in the application and validation of machine learning and artificial intelligence techniques. Contact him at helge@simula.no.%
\end{IEEEbiography}%

\end{document}